\documentclass[journal]{IEEEtran}
\usepackage[numbers,sort&compress]{natbib}
\usepackage{amsmath,amssymb,amsfonts}
\usepackage{graphicx}
\usepackage{textcomp}
\usepackage{xcolor}
\usepackage{hyperref}
\usepackage{multirow}
\usepackage{CJKutf8}
\usepackage{algorithm}
\usepackage{algpseudocode}
\usepackage{subfig} 
\usepackage{multirow}
\usepackage{graphicx}
\usepackage{tikz}

\emergencystretch=\maxdimen
\begin{document}

\title{A Hybrid Autoencoder-Transformer Model for Robust Day-Ahead Electricity Price Forecasting under Extreme Conditions
\thanks{This work was supported by the Shenzhen 
Research Institute of Big Data (under the internal project No. J00220240006). (\textit{Corresponding author: Jianghua Wu}) }
}

\author{
    \IEEEauthorblockN{
         Boyan Tang\IEEEauthorrefmark{1}\IEEEauthorrefmark{2}, 
         Xuanhao Ren\IEEEauthorrefmark{1}, 
         Peng Xiao\IEEEauthorrefmark{1},
         Shunbo Lei\IEEEauthorrefmark{3},
         Xiaorong Sun\IEEEauthorrefmark{4},
         Jianghua Wu\IEEEauthorrefmark{2}
    }
    
    \IEEEauthorblockA{
        \IEEEauthorrefmark{1}Beijing Normal-Hong Kong Baptist University, Zhuhai, Guangdong, China  \\
        \IEEEauthorrefmark{2}Shenzhen Research Institute of Big Data, The Chinese University of Hong Kong, Shenzhen, Guangdong, China \\
        \IEEEauthorrefmark{3}School of Science and Engineering, The Chinese University of Hong Kong, Shenzhen, Guangdong, China \\
        \IEEEauthorrefmark{4}School of Electrical and Power Engineering, Hohai University, Nanjing, Jiangsu, China \\
    }
\IEEEauthorblockA{E-mails: \{
s230034047@mail.uic.edu.cn, s230031274@mail.uic.edu.cn, s230034040@mail.uic.edu.cn,
\\leishunbo@cuhk.edu.cn, xsun@hhu.edu.cn, wujianghua@sribd.cn\}}
    
}

\maketitle

\begin{abstract}
Accurate day-ahead electricity price forecasting (DAEPF) is critical for the efficient operation of power systems, but extreme condition and market anomalies pose significant challenges to existing forecasting methods. To overcome these challenges, this paper proposes a novel hybrid deep learning framework that integrates a Distilled Attention Transformer (DAT) model and an Autoencoder Self-regression Model (ASM). The DAT leverages a self-attention mechanism to dynamically assign higher weights to critical segments of historical data, effectively capturing both long-term trends and short-term fluctuations. Concurrently, the ASM employs unsupervised learning to detect and isolate anomalous patterns induced by extreme conditions, such as heavy rain, heat waves, or human festivals. Experiments on datasets sampled from California and Shandong Province demonstrate that our framework significantly outperforms state-of-the-art methods in prediction accuracy, robustness, and computational efficiency. Our framework thus holds promise for enhancing grid resilience and optimizing market operations in future power systems.
\end{abstract}


\footnotetext[1]{For the emotional support function $f(Halcyon Yang)$ that always converges to happiness. (Corresponding author:TangBoyan)}

\section{Introduction}
Day-ahead electricity price forecasting (DAEPF) is vital to modern power system operations, providing important information for generators, market operators, and consumers. An accurate forecast helps generators schedule output, market operators formulate dispatch policies, and consumers plan energy usage and expenditure. In contrast, inaccurate forecasting threatens the stability of the whole market by triggering extreme price spikes or even negative electricity prices. Such danger has increased as electricity market regulations have increasingly been relaxed, allowing price signals to respond more strongly to variations in supply, demand, and extraneous factors. Recent events, e.g., record-breaking winter storms in parts of the United States and negative pricing events in Shandong province, China, have shown that extreme condition and other exceptional circumstances can increase price volatility, making improved forecasting capability a pressing necessity.

DAEPF has been extensively studied using both statistical and machine learning approaches. Traditional statistical methods, such as Autoregressive (AR) \cite{fosso1999}, Autoregressive Integrated Moving Average (ARIMA) \cite{li2007, zhang2021, huang2006, li2016}, and Generalized Autoregressive Conditional Heteroskedasticity (GARCH) \cite{garcia2005}, describe past patterns through time-series analysis. Although these methods effectively describe conventional price behavior, they typically depend on stationarity and normality assumptions that often fail under extreme market conditions. Moreover, they lack the flexibility to adapt to evolving dynamics, e.g., sudden policy changes or seasonal effects. Extensions like regime-switching models \cite{amjady2006} and wavelet-based decompositions \cite{zhang2005} have improved peak price accuracy. They, however, still perform inadequately when dealing with data exhibiting price spikes, heavy tails, and other complex behaviors.

Encouraged by the success of neural networks in capturing non-linear relationships, machine learning methods began to gain popularity in electricity price forecasting in the 1990s. Early efforts were based on feedforward neural networks, e.g., multilayer perceptrons (MLP), and backpropagation-based models \cite{lecun1998}. Subsequent research introduced more advanced network architectures, including convolutional neural networks (CNN) \cite{krizhevsky2012} and recurrent architectures such as long short-term memory (LSTM) \cite{szkuta1999} and gated recurrent units (GRU) \cite{bunn2000}. These methods better capture temporal dependencies in electricity prices and tend to perform better than traditional time-series models. Ensemble methods, which combine multiple learning algorithms, have also been found to be effective in improving forecasting performance \cite{nogales2002,shahidehpour2002}. However, deep networks often become inefficient for capturing very long-term dependencies due to issues like exploding or vanishing gradients and high computational demands, rendering them less suitable for 24-hour day-ahead predictions or volatile market conditions. Consequently, researchers have recently adopted attention mechanisms and encoder-decoder architectures for DAEPF \cite{kharlova2020,dong2021}. Such methods concentrate computational effort on the most informative parts of a sequence and better capture both local variations and overall market trends. Nevertheless, deep encoder-decoder models may still suffer from inefficiencies when processing large datasets or balancing short-term volatility with long-term policy impacts.

To this end, this paper introduces a novel DAEPF framework built upon the Informer architecture  \cite{zhou2021}, which incorporates the following two key innovations: 

\textbf{1. Distilled Attention Transformer (DAT).}
Our method introduces a DAT model. DAT automatically identifies and emphasizes the most influential segments of historical data, enabling it to balance long-term trends with short-term fluctuations effectively.

\textbf{2. Autoencoder Self-regression Model (ASM).}
To further improve the robustness under extreme conditions, our method integrates an Autoencoder self-regression model (ASM). ASM automatically recognizes and isolates atypical data patterns, such as those caused by heavy rain, heat waves, holidays, or massive human activities.

The incorporation of the above two innovations contributes to superior DAEPF performance for our method. In particular, DAT model dynamically assigns higher weights to critical time steps, e.g., those influenced by sudden weather changes or unexpected events, thereby enhancing the model's sensitivity to short-term anomalies. Concurrently, the ASM effectively detects and isolates these atypical patterns, ensuring that extreme events are modeled separately and do not distort the overall trend. This approach facilitates a balanced capture of long-term and short-term dependencies. Moreover, the streamlined computation inherent in the DAT model reduces the computational burden during real-time forecasting, while the modular architecture of the ASM decouples abnormal data processing from the primary forecasting task, thereby enhancing overall computational efficiency and responsiveness. We tested our model through experiments on datasets sampled from the California and Shandong Province markets,  confirming that our framework consistently outperforms state-of-the-art baselines in terms of accuracy, responsiveness, and computational efficiency. 

The rest of this paper is organized as follows. Section \ref{sec:II} defines and formulates the DAEPF problem. Section \ref{sec:III} outlines the entire process for solving the DAEPF problem with our proposed framework. Section \ref{sec:IV} provides a detailed description of our proposed architecture, including its DAT model and ASM process. Section \ref{sec:V} presents experimental results on several datasets, including ablation studies and comparisons with state-of-the-art algorithms. Section \ref{sec:VI} discusses key findings and practical implications for DAEPF including conclusion and future works.

\section{Problem Definition}
\label{sec:II}

This section presents the definition and the formulation of DAEPF problem and introduces two corresponding forecasting strategies, including single-step and iterative multi-step prediction strategies. The latter recursively utilizes previous predictions as inputs for forecasting subsequent prices. 

To formally define the DAEPF problem, let \textit{X} denote the historical electricity price, and is given as:

\begin{equation}
X = \{ x_1, x_2, \dots, x_T \} \
\end{equation}
where \( x_t \in \mathbb{R}^d \) represents the electricity price and related features (e.g., weather, load, etc.) at time step \( t \).

The goal of DAEPF problem is to predict the unknown electricity price \textit{Y}  for a future time period \( H \):
\begin{equation}
    Y = \{ y_{T+1}, y_{T+2}, \dots, y_{T+H} \}
\end{equation}
where \( y_{T+h} \in \mathbb{R} \) denotes the electricity price at time step \( T+h \). 

However, particularly, under extreme conditions, electricity price data may exhibit significant anomalous fluctuations, thus requiring a robust forecasting framework to handle both normal and extreme conditions.

In some cases, the forecasting task may be performed iteratively, where the predicted result at the current time step is used as input to the next time step to recursively predict the electricity price at subsequent time steps. For example, after predicting \( \hat{y}_{T+1} \), it is used as part of the input to predict \( \hat{y}_{T+2} \), and so on, until \( H \) steps of predictions are completed.

The forecasting framework is normally divided into two key strategies, i.e., single-step prediction and iterative multi-step prediction. 

\subsection{Single-Step Prediction}

In single-step prediction, the model leverages historical data to directly forecast the electricity price for the immediate next time step, capturing the immediate trends without incorporating any previous forecasts.

For time step \( T+1 \), use historical data \( X \) to predict:
\begin{equation}
    \hat{y}_{T+1} = f(X; \theta)
\end{equation}
where \( f(\cdot) \) is the forecasting model (e.g., Transformer) and \( \theta \) is the model parameters.

\subsection{Iterative Multi-Step Prediction}

In iterative multi-step prediction, the model recursively utilizes its previous predictions along with historical data to forecast future time steps, thereby extending the forecast horizon while continuously updating its inputs with the latest forecasted values.

For time step \( T+h \) (\( h \geq 2 \)), use historical data \( X \) and the predicted results from previous time steps \( \hat{y}_{T+1}, \hat{y}_{T+2}, \dots, \hat{y}_{T+h-1} \) to predict:
\begin{equation}
    \hat{y}_{T+h} = f\left( X \cup \{ \hat{y}_{T+1}, \hat{y}_{T+2}, \dots, \hat{y}_{T+h-1} \}; \theta \right)
\end{equation}

In this way, the \( H \)-step prediction is completed iteratively.

\section{Process Overview for Solving DAEPF}
\label{sec:III}

This section presents the complete workflow for addressing the DAEPF problem using our proposed framework. The process encompasses several key stages, including data preprocessing, model training, anomaly detection, data extraction, and extreme condition model training. Noted that our approach leverages two novel deep learning models, i.e., ASM and DAT, and the detailed descriptions and analyses of ASM and DAT will be provided in Section \ref{sec:IV}. Key steps of our process are outlined below.

\textbf{Data Preprocessing:} To enhance the model’s learning and generalization capabilities, the original data is first standardized using the following formula:
\begin{equation}
    X_{\text{standardized}} = \frac{X - \mu}{\sigma}
\end{equation}
where \( X \) represents the original data, \( \mu \) is the mean of the data, and \( \sigma \) is the standard deviation.

\textbf{Model Training:} The standardized data is fed into the Distilled Attention Transformer (DAT) model for conventional price forecasting. Simultaneously, the encoder and decoder components of the Autoencoder Self-regression Model (ASM) are trained. The training process minimizes the Mean Squared Error (MSE) between the predicted and actual prices:
\begin{equation}
    \mathcal{L}_{\text{MSE}} = \frac{1}{N} \sum_{i=1}^{N} (y_i - \hat{y}_i)^2
\end{equation}
where \( y_i \) is the true price, \( \hat{y}_i \) is the predicted price, and \( N \) is the number of data points.

\textbf{Anomaly Detection and Data Extraction:} The trained ASM is used to identify poorly fitted data points by calculating the reconstruction error \( e_i \) for each data point \( x_i \):
\begin{equation}
    e_i = \| x_i - \hat{x}_i \|_2^2
\end{equation}
where \( \hat{x}_i \) is the reconstructed data point. If the error exceeds a threshold, the point is considered an anomaly. Extreme condition data (usually 50 to 150 time steps) around these poorly fitted data points are then extracted and saved.

\textbf{Extreme Condition Model Training:} The extreme condition data extracted in the previous step is used to train a specialized variant of the DAT model that is tailored for handling extreme conditions. During forecasting, the pre-trained ASM is first used to detect anomalies in each prediction iteration. If no anomalies are detected, the conventional DAT model is used to generate the forecast for the current time step. When anomalies are detected, the extreme condition model is activated, and its output is combined with that of the conventional model to produce the final forecast.

\section{Innovative ASM and DAT models}
\label{sec:IV}

This section details the novel deep learning models that form the backbone of our forecasting framework for DAEPF under extreme conditions. Two primary models, i.e., ASM and DAT, are presented in subsections \ref{sec:IVA} and \ref{sec:IVB}, respectively. Subsection \ref{sec:IVC} describes the Self-Attention Distillation Mechanism, which invites convolutional pooling operations to reduce computational costs and memory requirements. In the subsection \ref{sec:IVD}, Generative Decoder is introduced to generate predictions for all time steps in parallel, significantly improving inference speed by eliminating the sequential dependency.

\begin{figure}[!ht]
\centering
\includegraphics[width=0.5\textwidth]{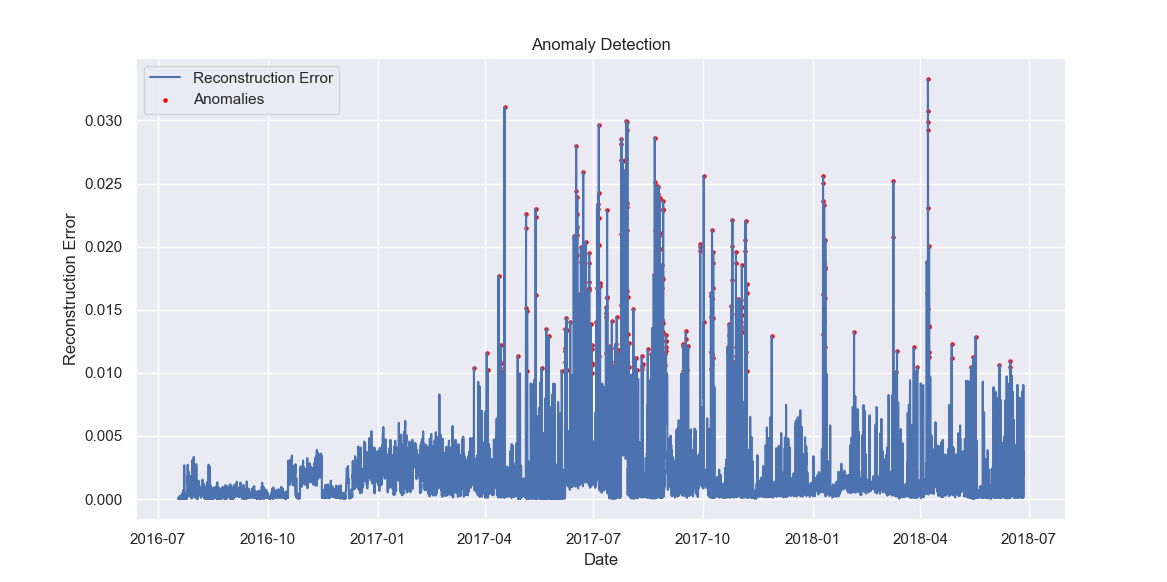}
\caption{Forecast Value with ASM Outliers (ETT-H-1 dataset)}
\label{fig:figure1}
\end{figure}

\subsection{Autoencoder Self-Regression Model (ASM)}
\label{sec:IVA}

Human festivals and extreme weather significantly disrupt electricity demand, posing forecasting challenges, especially for long-term and short-term predictions during extreme events (50-150 time steps). To address this, we introduce the Autoencoder Self-Regression Model (ASM), which isolates and analyzes extreme data through heavy regularization and specialized training, improving adaptability to extreme conditions.

An ASM uses unsupervised learning for dimensionality reduction and feature extraction. It has two parts: an encoder compressing data to a low - dimensional form and a decoder reconstructing the original data. The key goal is to minimize the reconstruction error between input and output. 

An ASM leverages unsupervised learning for dimensionality reduction and feature extraction. It consists of two parts: an encoder that compresses the data into a low-dimensional representation, and a decoder that reconstructs the original data. The core objective is to minimize the reconstruction error between the input and the output.

ASM excels at modeling complex data with outliers under extreme conditions, which traditional transformer-based models often fail to capture. Its unique design enables efficient extraction of extreme data patterns and effective anomaly detection. By compressing and reconstructing data, ASM identifies deviations through higher reconstruction errors. As an unsupervised method, it requires only normal data for training, simplifying data preparation and reducing costs.

After data cleaning, our algorithm identifies extreme data within the ASM framework. For selected extreme events, the inflection point is determined by averaging event data to create a representative time series of approximately  \( n \)  steps. Data from the first \( 2n \) steps is used to extract extreme-condition samples and build a specialized training set.

Using data from the preceding  \( 2n \)  steps, the model is trained to predict the next  \( n \)  extreme-value steps, capturing the dynamic evolution of extreme events for future predictions. When anomalies are detected, the system activates the extreme-condition model, improving robustness and accuracy in volatile environments.

\textbf{Encoder:} The encoder maps the input data \( x \) to the potential space representing \( z \). A common form of encoder is a neural network. 

\begin{equation}
z = f_{\theta}(x)
\end{equation}

\textbf{Decoder:} The decoder reconstructs the underlying representation \( z \) into the output \( \hat{x} \).

\begin{equation}
\hat{x} = g_{\phi}(z)
\end{equation}

\textbf{Loss function:} The goal of ASM is to minimize the difference between the input data \( x \) and the reconstructed data \( \hat{x} \). 

The training goal of ASM is to minimize the loss function \( \mathcal{L} \) by optimizing the parameters of the encoder and decoder \( \theta_e \) and \( \theta_d \).

\begin{equation}
\min_{\theta_e, \theta_d} L(x, g_{\theta_d}(f_{\theta_e}(x)))
\end{equation}
where the input data can be time-series data, images, or other types of data. The input is transformed into a low-dimensional latent representation \( z \) through an encoder function \( f_{\theta_e} \), typically implemented as a neural network. The latent representation captures the essential features of the input data in a compressed form. Subsequently, the decoder function \( g_{\theta_d} \) reconstructs the original data from the latent representation, generating the reconstructed output \( \hat{x} \).

\textbf{Reconstruction Error:} The reconstruction error for a given data point \( x \) is computed as the squared difference between the input and the reconstructed data:

\begin{equation}
e_i = \| x_i - \hat{x}_i \|^2
\end{equation}

\textbf{Anomaly Detection:} The anomaly score for data point \( x_i \) is defined by the reconstruction error \( e_i \). If the reconstruction error exceeds a predefined threshold \( \epsilon \), i.e., \( e_i > \epsilon \), the data point is then classified as an anomaly. Threshold \( \epsilon \) is normally determined based on the distribution of the reconstruction errors of normal data points. As shown in Figure \ref{fig:figure1}, it is the outlier identification data of the ETTh-1 dataset.

\subsection{Distilled Attention Transformer (DAT)}
\label{sec:IVB}

A key challenge in time-series forecasting is enhancing model receptive field and processing power without increasing computational costs.

We compared the time and memory of training and testing time of four model including Our Model, Transformer, LogTrans and LSTM (Table \ref{tab:table1}). Although LSTM has a low overhead for large-scale data processing O(\(L\)), its ability to handle complex time series cases is weak. Although the testing time of LogSparse Transformer (LogTrans) O(1) is faster than our model O(\(LlogL\)), the efficiency and accuracy of LogTrans are not as good. Transformers excel in handling time-series data, however, facing complexity due to the input-output asymmetry: longer input sequences (e.g., over 100 steps) provide more features but incur higher computational  O(\(L^2\)), exacerbated by their exponential complexity. Therefore, Our model improves the acceptance domain and processing power of the model without increasing the computational cost O(\(LlogL\)).

\begin{table}[ht]
\centering
\caption{Complexity Comparison}
\label{tab:table1}
\begin{tabular}{|c|c|c|c|}
\hline
\multirow{2}{*}{\textbf{Methods}} & \multicolumn{2}{c|}{\textbf{Training}} & \multirow{2}{*}{\textbf{Testing}} \\ \cline{2-3} 
                                  & \textbf{Time} & \textbf{Memory} & \textbf{Time} \\ \hline
Our Model                         & O(\(LlogL\))    & O(\(LlogL\))      & O(\(LlogL\))    \\ \hline
Transformer                       & O(\(L^2\))    & O(\(L^2\))      & O(\(L^2\))    \\ \hline
LogTrans                          & O(\(LlogL\))    & O(\(LlogL\))      & O(1)    \\ \hline
LSTM                              & O(\(L\))          & O(\(L\))            & O(\(L\))          \\ \hline
\end{tabular}
\end{table}


Transformers excel in NLP due to their self-attention mechanism, which replaces traditional sequence processing methods like CNNs and RNNs with attention and feedforward networks. This has made Transformer-based models, such as Informer, mainstream for tasks like time series forecasting.

However, traditional Transformers face challenges: self-attention computations involve dot-product operations, leading to exponential computational and memory costs as input sequences grow, limiting scalability. Additionally, stacking encoder and decoder layers for long sequences further increases memory demands, complicating long-sequence handling.

The original Transformer's dynamic decoding approach, where each step depends on prior predictions, slows down inference and incurs high computational overhead, particularly in real-time or batch processing scenarios.

To address these challenges, we propose an enhanced Transformer model with three key innovations: 

1. An Importance Self-Attention Mechanism that prioritizes critical time slots, reducing redundancy; 

2. Self-Attention Distillation \ref{sec:IVC}, which progressively shortens sequences per layer, improves efficiency;

3. A Generative Decoder that predicts \ref{sec:IVD} all time steps in parallel, eliminates autoregressive bottlenecks and speeds up predictions.

Our enhanced Transformer model was applied to long-term sequence forecasting, focusing on electricity generation and price prediction data from Shandong, China, and California, USA. Results show it outperforms traditional methods in accuracy and computational efficiency. In Figure \ref{fig:figure2}, the basic framework of our model is briefly revealed.


\subsection{Self-Attention Distillation Mechanism}
\label{sec:IVC}

Self-Attention Distillation is implemented by adding convolutional pooling between attention blocks, halving the sequence length at each layer to reduce input size, computation, and memory usage.

Assume that the length of the input is \( L \), by reducing the length into half at every layer, after the \( k \)-th layer the sequence length will be:

\begin{equation}
L_k = \frac{L}{2^k}
\end{equation}
where \( L_k \) is the sequence length after the \( k \)-th layer.

A given input matrix \( X \) of length \( L \) with the dimension \( d \) undergoes a convolutional pooling that has a new matrix length, which is reduced by one-half \( \frac{L}{2} \). In each layer of self-attention calculation over token similarities, the next down-sampling generates feature representations. This process will have to be repeated in layers of self-attention where at each subsequent layer of entry, the sequence is continually downsampled from the previously entering sequence.

After self-attention distillation, the sequence length is reduced by half at each layer, reducing the spatial complexity by a large margin. Suppose the length of an input sequence is \( L \):
\begin{itemize}
    \item First Layer after: Spatial Complexity = \( O(L \cdot d) \)
    \item After the second layer: Spatial Complexity = \( O\left(\frac{L}{2} \cdot d\right) \)
    \item After third layer: Spatial Complexity = \( O\left(\frac{L}{4} \cdot d\right) \)
\end{itemize}

Hence, after the \( k \)-th layer, spatial complexity is:

\begin{equation}
    O(L_k \cdot d) = O\left(\frac{L}{2^k} \cdot d\right)
\end{equation}

The objective of the self-attention distillation mechanism is to minimize the reconstruction error, that is, the difference between the input data \( x \) and the reconstructed data \( \hat{x} \). Our loss function is the mean squared error (MSE):

\begin{equation}
L(x, \hat{x}) = \| x - \hat{x} \|^2
\end{equation}

\begin{figure}[!ht]
\centering
\includegraphics[width=0.45\textwidth]{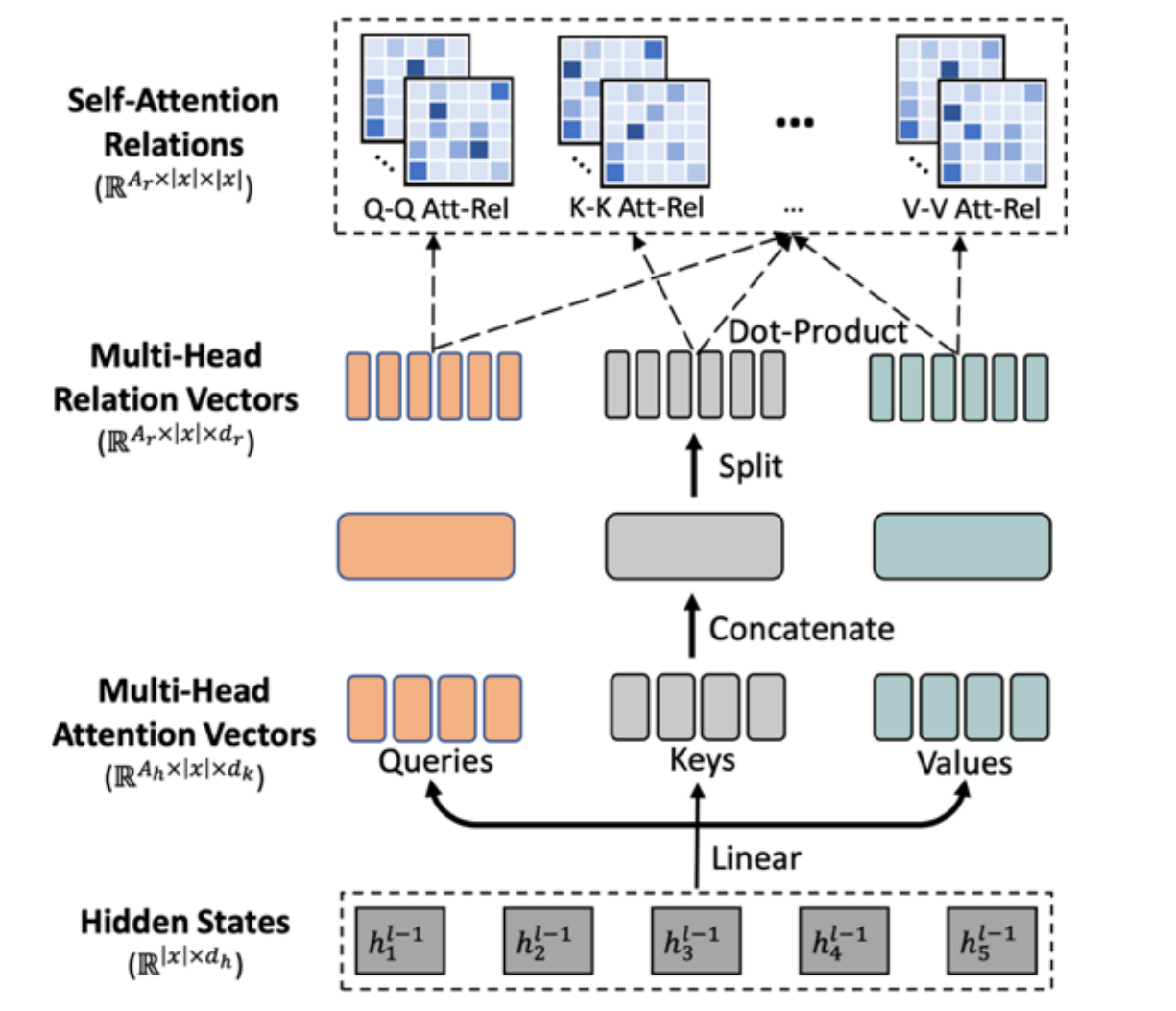}
\caption{Self-Attention Distillation Mechanism}
\label{fig:figure2}
\end{figure}

Through Self-Attention Distillation Mechanism (Figure \ref{fig:figure2}), it reduces computational load and memory consumption by shortening sequences, enabling efficient processing of long-sequence data. This technique effectively overcomes computational and memory bottlenecks in long-term time series forecasting, where traditional Transformers struggle.

\subsection{Generative Decoder}
\label{sec:IVD}

In traditional Transformer decoders, the prediction of each time step depends on the output of the previous time step. To speed up the prediction, we introduced the Generative Decoder, which eliminates traditional step-by-step inference by generating the outputs of all time steps at once. This reduces the time complexity from \( O(N) \) to \( O(1) \).

\textbf{Traditional Decoder:}
\begin{equation}
    \hat{y_t} = \text{Decoder}(\text{Encoder Output}, [y_1, y_2, \dots, y_{t-1}])
\end{equation}

\textbf{Generative Decoder:}
\begin{equation}
\hat{Y} = \text{Generative Decoder}(\text{Encoder Output})
\end{equation}

The generative decoder replaces the traditional step-by-step inference method and instead generates predictions for all time steps in one pass. In this approach, the decoder no longer depends on the output of the previous time step to generate the next. Instead, it considers the context of the whole sequence at once to generate all the predictions in one go.

\subsubsection{Components of Generative Decoder}

The output from the encoder is fed to the decoder: The generative decoder takes the output of the encoder, utilizing multi-layer self-attention mechanisms and feedforward networks to generate global encoded representations. These global representations capture the global dependencies and contextual information of the input sequence.

\subsubsection{All-time steps prediction}

Using all the context from the encoding, a generative decoder predicts a whole time sequence in one go, without going step-by-step in a process, achieving by the following formula:

\begin{equation}
\mathbf{Y} = \text{FeedForward}(\text{MultiHeadAttention}(\mathbf{X}_{\text{dec}}, \mathbf{H}_{\text{enc}}, \mathbf{H}))
\end{equation}
where \(\mathbf{X}_{\text{dec}}\) is the placeholder for the target sequence. \(\mathbf{H}_{\text{enc}}\) is the output of the encoder. \(\text{MultiHeadAttention}\) captures the global context information. \(\text{FeedForward}\) maps the global context to the final predictions \(\mathbf{Y}\). \(\mathbf{Y}\) represents the predictions for all time steps.

By producing the full output in one go with feed-forward networks, the generative decoder significantly improves computational efficiency and avoids the bottleneck of autoregressive inference.

\subsubsection{Parallelization and Time Complexity}

The generative decoder predicts all timesteps in parallel, eliminating the \( O(N) \)  bottleneck of traditional recursive methods and achieving constant-time \( O(1) \) predictions, drastically optimizing performance.

Our framework maps raw electricity data to a higher-dimensional space via embedding and positional encoding. Self-Attention Distillation extracts key features and shortens sequences, while the Generative Decoder predicts all steps in one pass, avoiding autoregressive bottlenecks. This resolves traditional Transformers' computational and memory issues, enabling robust long-term forecasting for energy and related industries.

\section{Case Evaluation}
\label{sec:V}

Our method has been implemented by using PyTorch CUDA 11.8 and Python 3.8. Testing was conducted on a Windows platform with an Intel i9-12900H processor, 16 GB RAM, 14 cores, and an RTX 3070 Ti GPU with CUDA 11.8. Three examples were trained and tested on multiple price datasets from (1) the 2012 California market, (2) the Shandong province market that obtained through the XMO platform \cite{xmo-opt}, and (3) a public benchmark dataset that called ETT dataset.

\subsection{Example 1: Extreme Conditions in the 2012 California Electricity Market}
To evaluate the model's predictive performance under extreme conditions, we utilized the 2012 California power market dataset, which captures the impact of severe heatwaves and sudden cold spells on power systems, electricity generation, and demand. This dataset provides an ideal testbed for assessing the model's robustness and adaptability in challenging scenarios.

To ensure a fair comparison across all models, we carefully selected test set data with strong periodicity and reliability. During result visualization, we applied smoothing techniques to minimize noise and enhance clarity, making the charts more intuitive. However, during the training phase, we only standardized the entire training dataset and avoided further cleaning or removal of noise or questionable data to maintain the integrity of the original data distribution.

\begin{table}[htbp]
\centering
\caption{Part of Ablation Experiment}
\label{tab:table2}
\begin{tabular}{|c|c|c|}
\hline
\textbf{} & \textbf{Our Model} & \textbf{Our Model*} \\ \hline
MSE       & 0.22               & 0.28                \\ \hline
MAE & 0.11        & 0.26                \\ \hline
\end{tabular}
\end{table}

To validate the role of the Auto-Encoder Self Regression Mechanism (ASM) in long-term sequence forecasting, we first conducted an ablation study. By comparing the performance of models trained on extreme data conditions versus those trained on the full dataset, we assessed the effectiveness of ASM. Specifically, we used Mean Squared Error (MSE) and Mean Absolute Error (MAE) to evaluate prediction accuracy and error control. As shown in Table \ref{tab:table2}, the model with ASM (MSE = 0.22, MAE = 0.11) significantly outperformed the model without ASM (MSE = 0.28, MAE = 0.26) in terms of prediction accuracy and error reduction. Additionally, the model trained solely on extreme data achieved better performance on most metrics compared to the model trained on the full dataset, demonstrating lower MSE and maximum absolute error. This indicates higher accuracy and robustness under extreme conditions.

\begin{figure}[t]
    \centering
    \subfloat{
    \includegraphics[width=0.4\textwidth]{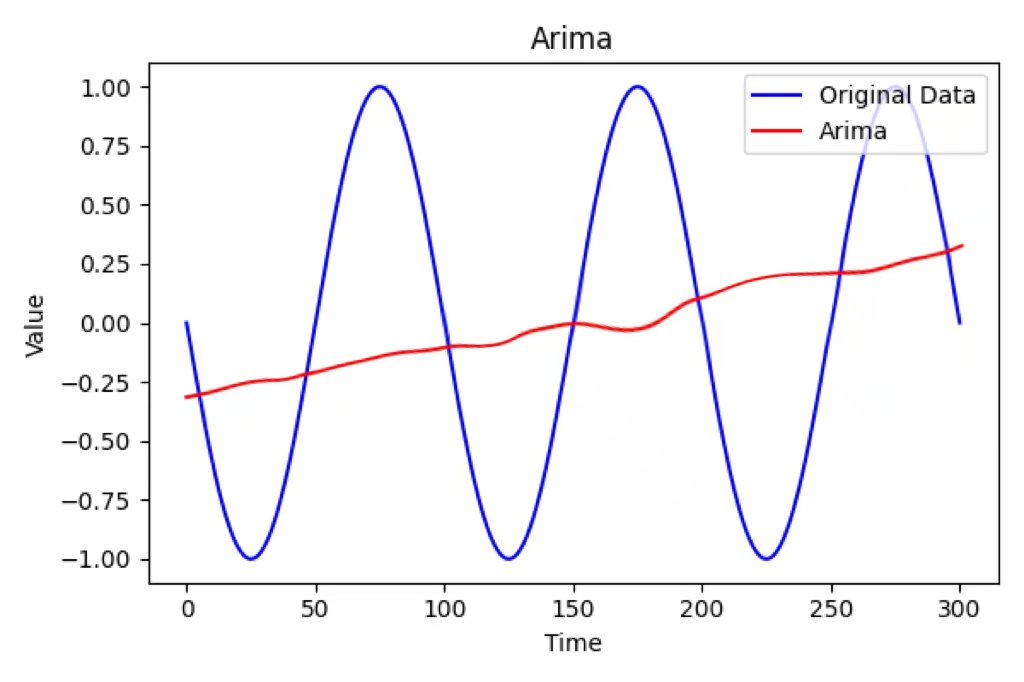}}

    \subfloat{
    \includegraphics[width=0.43\textwidth]{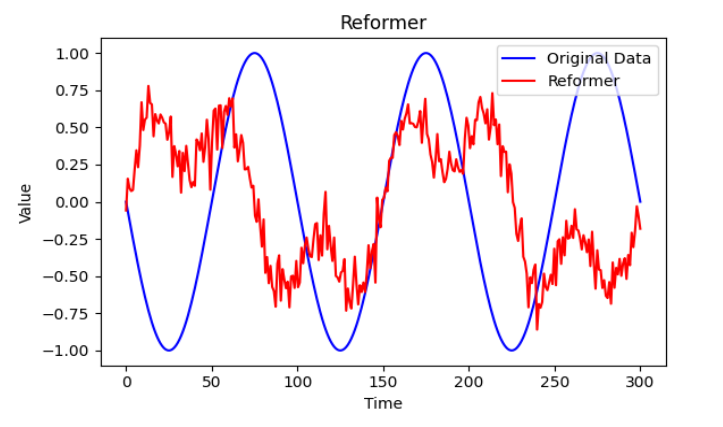}}
    
    \subfloat{
    \includegraphics[width=0.4\textwidth]{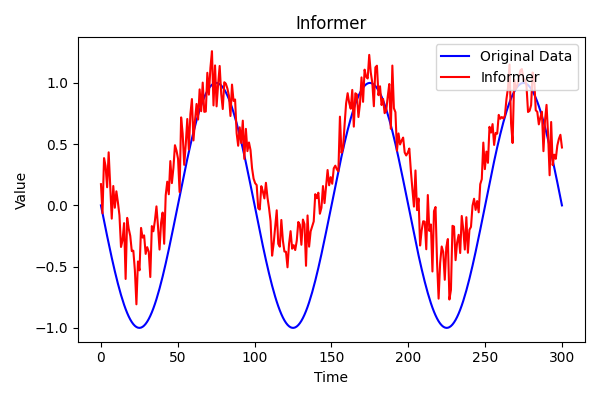}}
    
    \subfloat{
    \includegraphics[width=0.4\textwidth]{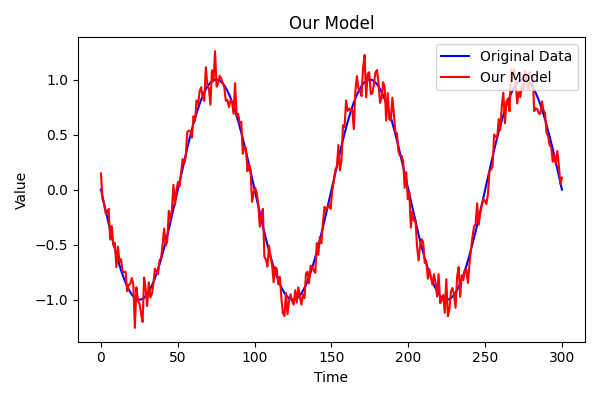}}
    \caption{Extreme Conditions Experiments}
    \label{fig:figure3}
\end{figure}

After validating the effectiveness of ASM, we further tested the model's predictive performance under extreme conditions. Using the 2012 California power generation dataset, we compared the performance of ARIMA, Informer, and our model. The experimental results (Figure \ref{fig:figure3}) show that ARIMA performed poorly under extreme conditions, while Informer struggled to accurately predict extreme points despite fitting most of the data well. In contrast, our model excelled in both regular and extreme conditions, accurately capturing electricity load trends and maintaining high stability and adaptability during sudden condition changes.

\subsection{Example 2: Generalization to Negative Prices in Shandong  Market}
To evaluate the model's generalization capability, we tested it on Shandong Province's electricity market data. Shandong's electricity prices are dynamically determined by supply and demand, presenting unique challenges such as complex price fluctuations, including the first recorded instance of negative electricity prices in China. The dataset captures significant pricing changes driven by market dynamics, such as sharp nighttime price drops due to low demand, as well as frequent fluctuations caused by extreme weather, holidays, and peak demand periods. These characteristics make the dataset highly nonlinear and uncertain, providing an ideal testbed for assessing model robustness.

\begin{table}[htbp]
\centering
\caption{Generalization Ability Evaluation}
\label{tab:table3}
\begin{tabular}{|c|c|c|c|}
\hline
\textbf{} & \textbf{ARIMA} & \textbf{Informer} & \textbf{Our Model}\\ \hline
MSE       & 1.475               & 0.832       & 0.829         \\ \hline
MAE & 1.285        & 0.698       & 0.672         \\ \hline
\end{tabular}
\end{table}

We compared our model to baseline methods, including ARIMA and Informer. Results are presented in Table \ref{tab:table3}), where shows that ARIMA achieved an MSE of 1.475 and an MAE of 1.285, while Informer recorded an MSE of 0.832 and an MAE of 0.698. Our model, superisely, achieved superior results with an MSE of 0.829 and an MAE of 0.672. From these numbers, we can find that traditional models like ARIMA struggle with dynamic time-series changes, often overfitting regular periods and performing poorly in volatile markets. Similarly, Informer exhibited higher errors during extreme weather and holidays, highlighting its limitations in handling complex market conditions. In contrast, our model demonstrated superior generalization and robustness, effectively capturing supply-demand-driven price fluctuations and adapting to complex market dynamics. It delivered more accurate price predictions and maintained stability even in highly volatile markets, with significantly lower errors than baseline methods.

\begin{table*}[htbp]
        \centering
        \caption{Model Comprehensiveness Evaluation}
        \includegraphics[width=0.7\textwidth, height=0.4\textwidth]{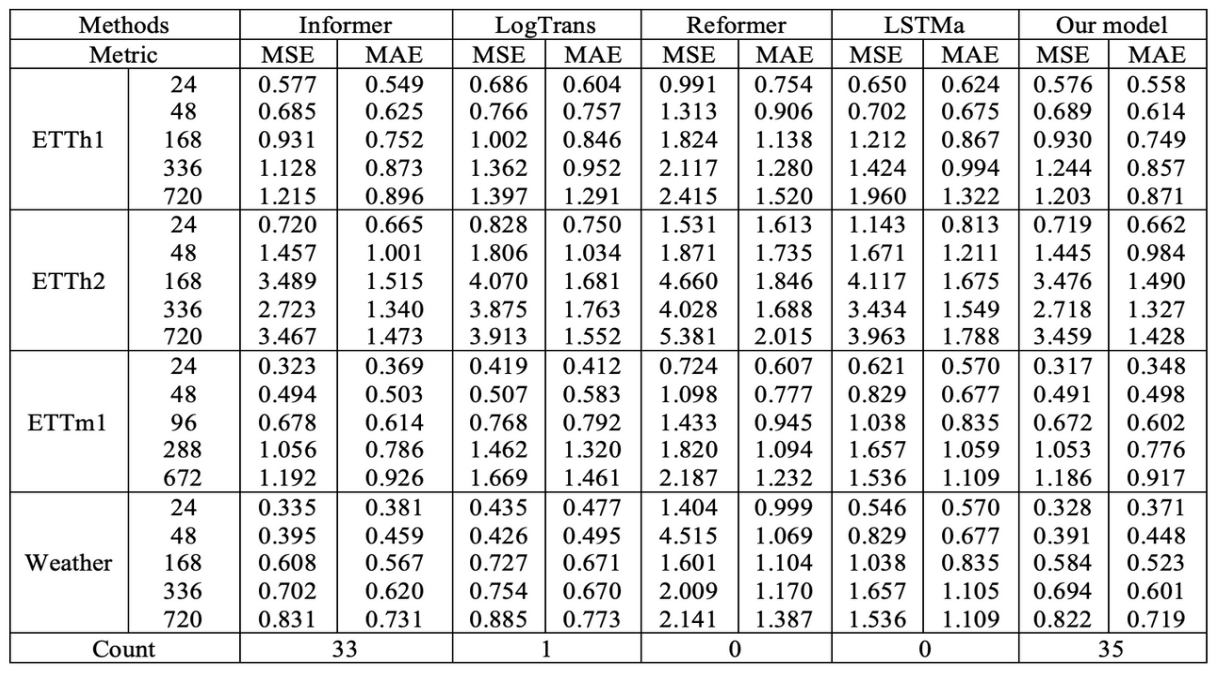}
        \label{fig:tableIV}
\end{table*}

\subsection{Example 3: Benchmark Performance on the ETT Dataset}
The model was further evaluated on the Electricity Transformer Temperature (ETT) dataset, a widely used benchmark for power system price forecasting. The ETT dataset includes three subsets: ETT-H-1, ETT-H-2, and ETT-M, each representing different time granularities and characteristics. ETT-H-1 and ETT-H-2 are hourly datasets recording transformer temperatures and power load, with ETT-H-2 covering a longer time period, making it suitable for long-term forecasting. Both subsets exhibit strong temporal dependencies and seasonal variations, testing the model's ability to handle periodic fluctuations and trends. In contrast, ETT-M is a high-frequency, minute-level dataset designed to evaluate the model's robustness and precision in complex, high-frequency scenarios. Training and prediction times varied across datasets. ETT-H-1 required approximately 2 hours, while ETT-M needed 3 hours. For comparison, the original Transformer implementation took about 6 hours to train on the ETT-M dataset, highlighting the efficiency of our approach. 

The test results are shown in Table \ref{fig:tableIV}. From the table, it is easy to find that our model achieves state-of-the-art (SOTA) performance on standard datasets. Based on MSE and MAE metrics, it competes with leading models in tasks such as energy consumption prediction using the ETT-H-1 dataset, long-term forecasting with ETT-H-2, and high-frequency classification on ETT-M. Importantly, our model excels in extreme conditions, maintaining stability with minimal performance degradation. While other models struggle with sudden data pattern shifts or anomalies caused by extreme weather or market fluctuations, our model remains resilient, delivering stable electricity price forecasts and significantly reducing the impact of extreme events.

As a preliminary work, there are several improvements can be made to enhance the performance. Currently, the model may not fully capture subtle fluctuations caused by sudden weather shifts or large-scale human activities. Additionally, the integration of exogenous factors, such as economic indicators, policy changes, and renewable energy patterns, remains an area for future exploration. Finally, incorporating real-time forecasting and online learning capabilities could further improve the model's ability to respond flexibly to sudden market changes or supply disruptions.


\section{Conclusion and Future Work}
\label{sec:VI}

This paper presented a hybrid deep learning framework for Day-Ahead Electricity Price Forecasting (DAEPF) under extreme conditions. Our approach integrates the Distilled Attention Transformer (DAT) with the Autoencoder Self-regression Model (ASM), and becomes effective in handling complex temporal dynamics and anomalies. The DAT efficiently captures key features across varying time scales, while the ASM accurately detects and isolates extreme data, leading to improved prediction accuracy and robustness. Future work will focus on improving anomaly detection, incorporating additional exogenous factors, and developing online continual learning capabilities to further enhance forecasting performance in dynamic market environments.


\bibliographystyle{IEEEtran}
\bibliography{IEEEabrv,mylib}

\end{document}